\documentclass[10pt, a4paper]{article}
\usepackage{lrec2016}
\usepackage{multibib}
\usepackage{diagbox}
\newcites{languageresource}{Language Resources}
\usepackage{graphicx}
\usepackage{mathtools}
\usepackage[b]{esvect}
\usepackage{pst-node}
\usepackage{arydshln}
\usepackage{hyperref}
\usepackage{epstopdf}

\title{\textbf{CLaC} @ QATS: \textbf{Q}uality \textbf{A}ssessment for \textbf{T}ext \textbf{S}implification}

\name{Elnaz Davoodi and Leila Kosseim}

\address{Concordia University \\
        Montreal, Quebec, Canada\\
         e\_davoo@encs.concordia.ca, kosseim@encs.concordia.ca\\}

\abstract{
This paper describes our approach to the 2016 QATS quality assessment shared task. We trained three independent Random Forest classifiers in order to assess the quality of the simplified texts in terms of grammaticality, meaning preservation and simplicity. We used the language model of Google-Ngram as feature to predict the grammaticality. Meaning preservation is predicted using two complementary approaches based on word embedding and WordNet synonyms. A wider range of features including TF-IDF, sentence length and frequency of cue phrases are used to evaluate the simplicity aspect. Overall, the accuracy of the system ranges from 33.33\% for the overall aspect to 58.73\% for grammaticality.\\ 
\newline 
 }

\begin{document}

\maketitleabstract
\section{Introduction}
%Quality assessment of text simplification is the task of judging how good a text is simplified regarding specific criteria including meaning preservation, simplicity level, etc. 
Automatic text simplification is the process of reducing the complexity of a text to make it more accessible to a broader range of readers with different readability levels. While preserving its meaning as much as possible, a text's lexical, syntactic and discourse level features should be made more simple. However, evaluating the simplicity level of a text is still a challenging task for both humans and automatic systems.

Current approaches to automate text simplification vary depending on the type of simplification. Lexical simplification was the first effort in this area. In particular, \newcite{devlin1998} introduced an approach of replacing words with their most common synonym based on frequency \cite{ku1967}. More recently, publicly available resources such as Simple English Wikipedia \footnote{\url{http://www.cs.pomona.edu/~dkauchak/simplification/}} and the Google 1T corpus \footnote{\url{https://books.google.com/ngrams}} have been used to automate lexical simplification based on similar approaches such as common synonym replacement and context vectors (e.g. \cite{biran2011,bott2012,rello2013,kauchak2013}).

Another approach to automatic text simplification involves syntactic simplification. Current work in this area aims to identify and simplify complex syntactic constructions such as passive phrases, embedded clauses, long sentences, etc. Initial work on syntactic simplification focused on the use of transformation rules in order to generate simpler sentences (e.g. \newcite{chandrasekar1997}). Later, work have paid more attention on sentence splitting (e.g. \newcite{carroll1998}), rearranging clauses (e.g. \newcite{siddharthan2006}) and dropping clauses (e.g. \cite{barlacchi2013,vstajner2013}). To our knowledge, \newcite{siddharthan2003} is the only effort that specifically addressed the preservation of a text's discourse structure by resolving anaphora and ordering sentence. 

%\textbf{more on evaluation of simplification}

In the remainder of this paper, we describe the methodology we used to measure the 4 simplification criteria of the QATS workshop: \textsc{Grammaticality}, \textsc{meaning preservation}, \textsc{simplicity} and \textsc{overall}. In Sections 2 and 3, the details of our submitted system are described, while Section 4 summarises our results. 

\iffalse
\section{Data and Evaluation}
This shared task focuses on automating text simplification from various aspects in order to predict: grammatical correctness of simplified sentences, degree of simplicity, meaning preservation and finally the combination of all. The task organizers released 505 pairs of original and simple sentences. The original sentence was taken from news domain and Wikipedia and the simple counterpart was automatically simplified using various text simplification systems. Thus, the simple counterparts may contain various types of simplifications such as lexical, syntactic or mixture of both. Table \ref{data} shows the distribution of the data for each of the four aspects. Our system aimed to predict the class-label of each of these four aspects.

\begin{table}

\begin{tabular}{l|c|c|c}\hline
\diagbox[width=3.5cm]{Aspect}{Value(\%)}&
  Good & Ok & Bad\\ \hline \hline
Grammaticality &  75.65 & 14.26 & 10.09 \\ \hline
Meaning preservation & 58.22 & 26.33 & 15.45\\ \hline
Simplicity & 52.68 & 30.29 & 17.03 \\ \hline
Overall & 26.33 & 46.14 & 27.53\\ \hline

\end{tabular}
\caption{Distribution of data}
\label{data}
\end{table}
\fi

\section{System Overview}
\begin{figure*}[!h]
    \centering
    \includegraphics[scale=0.6]{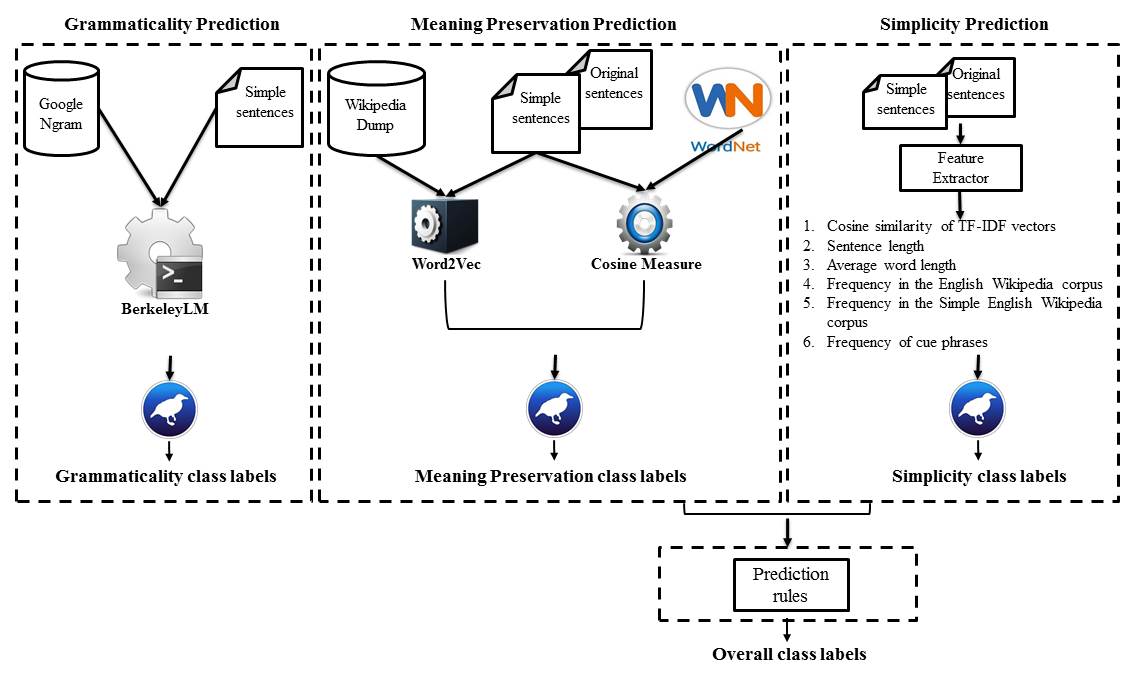}
    \caption{System Overview}
    \label{fig:overview}
\end{figure*}

As can be seen in Figure \ref{fig:overview}, our system consisted of three independent supervised models in order to predict each of the three main aspects: \textsc{grammaticality}, \textsc{meaning preservation} and \textsc{simplicity}. We used 10 fold cross-validation in order to choose the best supervised models. The $4^{th}$ aspect (i.e. \textsc{overall}) was predicted using the predictions of \textsc{meaning preservation} and \textsc{simplicity}. 

\subsection{Grammaticality Prediction}
In order to predict the quality of the simplified sentences from the point of view of grammaticality, we have used the log likelihood score of the sentences using the Google Ngram corpus\footnote{\url{https://books.google.com/ngrams}}. To do this, the BerkeleyLM language modeling toolkit\footnote{\url{http://code.google.com/p/berkeleylm/}} was used \cite{pauls2011} to built a language model from the Google Ngram corpus, then the perplexity of all simple sentences in the training set were calculated. These log likelihood scores were used as features to feed a Random Forest classifier.  

\subsection{Meaning Preservation Prediction}
The purpose of meaning preservation is to evaluate how close the meaning of the original sentence is with respect to its simple counterpart. To do this, we used two complementary approaches based on word embedding and the cosine measure.

\subsubsection{Word Embedding}
We used the Word2Vec package \cite{mikolov2013,mikolov2013-2} to learn the representation of words on the Wikipedia dump\footnote{\url{http://www.cs.pomona.edu/~dkauchak/simplification/}}. We then trained a skip-gram model using the \textit{deeplearing4j}\footnote{\url{http://deeplearning4j.org/}} library. As a result, each word in the original sentence and its simple counterpart are represented as a vector. As calculating the similarity of two sentences using word embedding is still a challenging task, our approach to this problem was to use average similarity. To do so, we calculated the similarity of each word (each vector) in the original sentence to all the words in its simpler counterpart. Then using the average length of the original and simple pairs, we calculated the average similarity between a pair of sentences. This similarity was the first feature we fed to a Random Forest classifier. 

\subsubsection{Cosine Similarity of WordNet Synonyms}
The second feature we used for meaning preservation was based on WordNet synonyms. Each sentence was represented as a vector of its constituent words. Then, using WordNet\footnote{\url{https://wordnet.princeton.edu/}}, all synonyms of each word were added to the corresponding vector of the sentence. However, as each word can have various part of speech (POS) tags, before expanding the vector, we first identified the POS of all the words in the sentence using the Stanford POS tagger \cite{manning2014}. Afterwards, we filtered the synonyms according to the POS tags and added only those with the same POS tag of the word. As a result, each sentence was represented as a vector of words and their synonyms. Using the cosine similarity to calculate the similarity between corresponding vectors of pairs of sentences, we measured how close the meaning of two sentences were. 

\begin{equation}
\label{cosinSim}
    %\[
    Cosine\_Sim(i) = \frac{\vv{O_{i}}. \vv{S_{i}}}{||\vv{O_{i}}||\times||\vv{S_{i}}||}
    %\]
\end{equation}

\subsection{Simplicity Prediction}
The purpose of simplicity prediction is to evaluate how simpler the simple sentences are compared to their original counterpart. As simplicity can be measured at various levels (i.e. lexical, syntactic and discourse), we considered the following sets of features in order to capture the changes at each level.

\subsubsection{Vector Space Model Similarity}
The first feature we considered in order to evaluate the simplicity of the simple sentences compared to their original counterpart, was the cosine similarity between the Term Frequecy-Inverse Document Frequency (TF-IDF) vectors of each pair. A cosine similarity of 1 indicates that no change has been made in the simplification process. However, before transforming sentences into their corresponding TF-IDF vector, we preprocessed them. First, stop words were removed, then all words were stemmed using the Porter Stemmer \cite{porter1980}. As a result, each sentence was represented as a vector of the size of all the stems in all sentences. It is worth noting that in order to compute the inverse document frequency for each stem, we considered each sentence as a document. The cosine similarity between original and simple sentences of the $i^{th}$ pair is calculated using Formula \ref{cosinSim}, where $\vv{O_{i}}$ and $\vv{S_{i}}$ represent the vectors of the original sentence and its simple counterpart correspondingly.

\subsubsection{Sentence Length}
Traditional approaches to readability level assessment have identified text length as an important feature to measure complexity (e.g. \newcite{kincaid1975}). Following this, we investigated the influence of sentence length in terms of the number of open class words only. By ignoring closed class words, we eliminated the effect of words which do not contribute much to the meaning of the sentence. Thus, we considered the difference between the length of pairs of sentences as our second feature for simplicity prediction. 

\subsubsection{Average Word Length}
According to \newcite{kincaid1975}, not only can the number of words in the sentence be an indicator of simplicity level, but also its length in terms of the number of characters. To account for this, we also considered the difference between the average number of characters between pairs of sentences. Using this feature along with the number of words of each sentence (see Section 2.3.2), we investigated not only the influence of the length of sentence, but also the length of each word in the sentence.

\subsubsection{Frequency in the English Wikipedia Corpus}
\label{EnglishWiki}
The frequency of each word in the regular English Wikipedia can be an indicator of the simplicity level of the word. We expected that words in the original sentences would be more frequent in the regular English Wikipedia than words of the simple sentences. Thus, we calculated the difference between the average frequency of all words of the original sentence and their simple counterpart. To do this, we preprocessed both pairs of sentences and the regular English Wikipedia corpus, in order to remove stop words and then stem the remaining words. 

\subsubsection{Frequency in the Simple English Wikipedia Corpus}
The Simple English Wikipedia corpus\footnote{\url{http://www.cs.pomona.edu/~dkauchak/simplification/}} is an aligned corpus of 60K \textit{$<$regular, simple$>$} pairs of Wikipedia articles. We used this corpus in order to calculate the average frequency of words of each pair of sentences. We expected the words of simpler sentences to be more frequent in the Simple Wikipedia articles compared to the original sentences. To do this, we performed the same preprocessing as described in Section \ref{EnglishWiki} and used the average frequency of the sentence's stems as features.

\subsubsection{Frequency of Cue Phrases}
The last feature we considered to predict the simplicity aspect was the difference in the usage of cue phrases. Cue phrases are special terms such as \textit{however, because, since, etc.} which connect text segments and mark their discourse purpose. Several inventories of cue phrases have been proposed (e.g. \cite{knott1996,prasad2007}). For our work, we used the list of 100 cue phrased introduced by \newcite{prasad2007} and calculated the difference between the frequency of cue phrases across pairs of sentences. It is worth noting that cue phrases may be used to explicitly signal discourse relations between text segments or may be used in a non-discourse context. However, here we considered both discourse and non-discourse usage of cue phrases.

\subsection{Overall Prediction}
The last aspect to be predicted evaluated the combination of all other aspects. According to our analysis of the training data set, this aspect depended mostly on the \textsc{simplicity} and the \textsc{meaning preservation} aspects. Our prediction of this aspect was based only on a simple set of rules using the predictions of these two aspects. The following shows the rules we used to predict the value of this aspect.

\begin{itemize}
    \item If \textbf{both} \textit{simplicity} and \textit{meaning preservation} are classified as \textsc{good}, then \textit{overall} $=$ \textsc{good},
    \item If \textbf{at least} one of \textit{simplicity} or \textit{meaning preservation} is classified as \textsc{bad}, then \textit{overall} $=$ \textsc{bad},
    \item otherwise, \textit{overall} $=$ \textsc{ok}.
\end{itemize}

\section{Data and Results}
The training set contains 505 pairs of original and simple sentences. The original sentences were taken from the news domain and from Wikipedia and the simple counterparts were automatically simplified using various text simplification systems. Thus, the simple counterparts may contain various types of simplifications such as lexical, syntactic or mixture of both. Table \ref{data} shows the distribution of the data for each of the four aspects. As can be seen, none of the aspects have a normal distribution over the class-labels. 

\begin{table}[h!]
\centering
\begin{tabular}{|l|c|c|c|}\hline
\diagbox[width=3.5cm]{Aspect}{Value(\%)}&
  Good & Ok & Bad\\ \hline \hline
Grammaticality &  75.65 & 14.26 & 10.09 \\ \hline
Meaning preservation & 58.22 & 26.33 & 15.45\\ \hline
Simplicity & 52.68 & 30.29 & 17.03 \\ \hline
Overall & 26.33 & 46.14 & 27.53\\ \hline
\end{tabular}
\caption{Distribution of data}
\label{data}
\end{table}

For our participation, we submitted one run for \textsc{Grammaticality} and \textsc{Meaning Preservation} and three runs for the \textsc{Simplicity} and \textsc{Overall} aspects. The three runs had different classification threshold to assign class labels.
Our official results are listed in Table \ref{Result}. MAE and RMSE stand for Mean Average Error and Root Mean Square Error correspondingly. 

\begin{table*}
\centering
\begin{tabular}{|l|c|c|c|}\hline
%\diagbox[width=3.5cm]{Aspect}{Value(\%)}&
  \bf System Name & \bf Accuracy & \bf MAE  & \bf RMSE \\\hline \hline
Grammaticality-Davoodi-RF-perplexity & 58.73\% & 27.38 & 34.66\\ \hdashline
Meaning preservation-Davoodi-RF & 49.21\% & 30.56 & 36.31\\\hdashline
Simplicity-Davoodi-RF-0.5 & 34.92\% & 43.25 & 45.32\\
Simplicity-Davoodi-RF-0.6 & 35.71\% & 41.67 & 44.48\\
Simplicity-Davoodi-RF-0.7 & 35.71\% & 40.48 & 44.48\\\hdashline
Overall-Davoodi-0.5& 34.13\% & 41.27 & 44.21 \\
Overall-Davoodi-0.6 & 32.54\% &	42.06 & 45.67\\
Overall-Davoodi-0.7 & 33.33\% & 41.27 & 45.16\\ \hline
\end{tabular}
\caption{Official Results of our system at QATS.}
\label{Result}
\end{table*}

\section{Bibliographical References}
\label{main:ref}

\bibliographystyle{lrec2016}
%\bibliographystyle{plain}
%\bibliographystyle{plainnat}
%\bibliography{xample}

\end{document}